\documentclass[lettersize,journal]{IEEEtran}
\usepackage{amsmath,amsfonts}
\usepackage{algorithmic}
\usepackage{array}
\usepackage[caption=false,font=normalsize,labelfont=sf,textfont=sf]{subfig}
\usepackage{textcomp}
\usepackage{stfloats}
\usepackage{url}
\usepackage{verbatim}
\usepackage{graphicx}
\usepackage{authblk}

\def\BibTeX{{\rm B\kern-.05em{\sc i\kern-.025em b}\kern-.08em
    T\kern-.1667em\lower.7ex\hbox{E}\kern-.125emX}}
\usepackage{balance}
\begin{document}
\title{Attentions Under the Microscope: A Comparative Study of Resource Utilization for Variants of Self-Attention}

\author[1,*]{Zhengyu Tian}
\author[1]{Anantha Padmanaban Krishna Kumar}
\author[1]{Hemant Krishnakumar}
\author[3]{Mehdi Hosseinzadeh}
\author[1,2,*]{Reza Rawassizadeh}
\affil[1]{Department of Computer Science, Boston University Metropolitan College}
\affil[2]{Center of Excellence in Precision Medicine and Digital Health, Department of Physiology, Chulalongkorn University, Thailand}
\affil[3]{School of Engineering \& Technology, Duy Tan University, Da Nang, Vietnam;}
\affil[*]{Corresponding authors: \texttt{zytian@bu.edu, rezar@bu.edu}}

\maketitle

\begin{abstract}
% where to submit: TMLR, Sustainable Computing: Informatics and Systems (SUSCOM), IEEE Transactions on Sustainable Computing (TSUSC) , IGSC (International Green and Sustainable Computing Conference)
Large language models (LLMs) and vision-language models (VLMs) are scaling rapidly and now underpin a wide range of applications. Nearly all are built on the transformer architecture, whose attention layers have become a central computational bottleneck because their time and memory cost grows quadratically with sequence length. Many efficient attention variants have been proposed to address this, yet their actual energy consumption and hardware resource demands during training remain poorly characterized. In this work, we benchmark eight attention mechanisms across the training of five transformer-based models, measuring training time, GPU memory, FLOPs, CPU utilization, and power draw.
We find that the most energy-efficient mechanisms---Flash Attention,
Locality-Sensitive Hashing (LSH) Attention, and Multi-Head Latent Attention (MLA), reduce energy through complementary routes: kernel-level, algorithmic, and architectural optimization, respectively. We further show that lower instantaneous GPU power does not by itself guarantee lower energy use, since training time contributes just as much to the total. Our energy-aware benchmark offers practical guidance for selecting resource-efficient attention mechanisms.
Code is available at \footnote{https://github.com/Zhengyu-Tian/attention-benchmark}.
\end{abstract}


\section{Introduction and Background}
Since the release of ChatGPT in late 2022, large language models (LLMs) and visual language models (VLMs) have become increasingly prevalent across a wide range of applications. On the other hand, clean water consumption, huge electricity usage, and massive heat emission of AI datacenter challenges global natural resources severely \cite{Morrisonetal2025, Ebertetal2024, Dooetal2024}.

A de-facto model to build LLMs and VLMs is the transformer architecture \cite{vaswani2017attention}. A transformer is composed of an encoder and decoder, and each of these parts contains two main neural network components, self-attention layers and feed-forward layers \cite{rezabook}. Both self-attention and feed-forward layers are extremely resource-intensive. Previous studies \cite{zhao2023survey, brown2020language, kaplan2020scaling} show that models with a larger number of parameters usually have higher accuracy than models with smaller numbers of parameters. As model sizes continue to grow, the resource utilization of transformers has become a major challenge, affecting global warming \cite{bux2025critical}, training efficiency \cite{wen2025grades}, and deployment of these models on resource-constrained devices, such as personal computers, or smartphones \cite{ejimuda2025, odsearch23}.

To reduce the overhead of feed-forward layers, approaches such as Low-Rank Adaptation (LoRA) \cite{hu2022lora} or Mixture of Experts (MoE) \cite{cai2024survey} have been proposed. Researchers have also developed various alternative or optimized self-attention mechanisms, such as Flash Attention \cite{dao2023flashattention2}, Linear Attention \cite{katharopoulos2020transformers}, Locality-Sensitive Hashing (LSH) Attention\cite{kitaev2020reformer}, and Sliding Window Attention \cite{beltagy2020longformer}, which theoretically offer low computational overhead. Although these attention mechanisms are often considered to be efficient or energy-saving in the literature, their actual performance in terms of hardware resource usage, training time, and energy consumption during training remains insufficiently evaluated \cite{tay2022efficient}. 
We believe, because of increasing global concern about the environmental impact of AI models, including drinking water shortage, electricity consumption and carbon footprint \cite{lacoste2019quantifying, patterson2021carbon, schwartz2020green, strubell2020energy}, the power consumption from model training and inference requires more quantitative analysis. Existing studies mostly focus on model accuracy or inference performance, while the energy cost during training has received little attention.

In this study, we evaluate eight different self-attention mechanisms through a unified benchmarking framework, and systematically measuring their resource usage of five different transformer-based models training. To benchmark the resource utilizations of these methods, we monitor seven indicators, including (i) training time, (ii) GPU memory usage, (iii) model size, (iv) GPU Floating Point Operations Per Second (FLOPS), (v) CPU and (vi) GPU utilization percentage, and (vii) GPU power consumption. 

Studying these parameters assists model builders in further development of transformer-based models and provides reference data for reducing the environmental challenges imposed by AI models. Our work does not directly measure carbon emissions, but rather supports future estimation efforts by providing energy consumption data that can be combined with region-specific electricity carbon intensity factors.

Our experimental results reveal significant differences in training efficiency and energy consumption across attention mechanisms. These findings offer practical guidance for selecting appropriate architectures in resource-constrained environments, in line with Green AI principles \cite{schwartz2020green}, which advocate for energy-efficient and environmentally sustainable AI development. Furthermore, our results provide quantitative reference data to help enterprises and developers make informed decisions when choosing attention mechanisms, considering training cost, and their available hardware infrastructure.

\section{Variants of Self-Attention}

The transformer architecture, introduced by Vaswani et al. \cite{vaswani2017attention}, is the foundation of most practical language and vision-language models  \cite{zhao2023survey}. Its self-attention mechanism enables the modeling of token dependencies across long sequences. However, the original attention formulation incurs quadratic complexity in both time and memory \cite{tay2022efficient}. This high cost not only consumes substantial computational resources but also increases operational expenses for enterprise deployment and poses potential environmental concerns due to energy demand. This led to the introduction of several optimizations on self-attention models.

\subsection{Self-Attention Variants}
To improve the efficiency of self-attention computation, many alternative designs have been proposed. Research has explored different modifications such as Grouped Query Attention \cite{ainslie2023gqa}, Sliding Window Attention \cite{beltagy2020longformer}, LSH Attention \cite{kitaev2020reformer}, and Flash Attention \cite{tay2022efficient}, which is optimized for NVIDIA GPU. These designs aim to reduce resource consumption while maintaining model performance. 

\begin{itemize}
    \item \textbf{Baseline self-Attention \cite{vaswani2017attention}:} The classic self-attention module as implemented in GPT-2, using masked scaled dot-product attention with full sequence visibility restricted to past tokens. It serves as the baseline for comparison in this study.
    
    \item \textbf{Scaled Dot-Product Attention \cite{luong2015effective}:} This mechanism computes attention scores by scaling and taking the  Dot-Product between queries and keys, followed by a softmax operation. It incurs $O(n^2)$ time and memory complexity.

    \item \textbf{Grouped Query Attention \cite{ainslie2023gqa}:} This mechanism groups queries into several partitions and computes attention separately within each group, reducing computational overhead and memory usage while maintaining sufficient representation capacity.

    \item \textbf{Linear Attention \cite{katharopoulos2020transformers}:} A variation of attention that applies a simple positive feature map (e.g., ReLU+1) to both queries and keys, enabling the approximation of traditional attention with linear time complexity while maintaining expressiveness. It has a hardware efficient variety such as lightning attention \cite{li2025minimax}, but we analyze the common linear attention, due to its generalizability. 

    \item \textbf{Sliding Window Attention \cite{beltagy2020longformer}:} Instead of attending to the full sequence, sliding window attention restricts attention computation to a fixed-size local window around each token, significantly reducing complexity for long sequences. 

    \item \textbf{Locality-Sensitive Hashing (LSH) Attention \cite{kitaev2020reformer}:}  This attention mechanism clusters similar tokens into the same buckets using the random rotation-based hash function \cite{charikar2002}, and performs attention only within buckets, reducing full sequence attention to sparse approximations.

    \item \textbf{Flash Attention version 2 \cite{dao2023flashattention2}:} A highly optimized attention implementation that reduces memory access overhead and improves computational efficiency by reorganizing Nvidia's GPU memory patterns and leveraging GPU parallelism, leading to faster training and lower memory consumption. It is one of the common attention mechanisms used in different LLMs, such as Llama \cite{touvron2023llama} and Mistral \cite{chaplot2023albert} model series.

    \item \textbf{Multi-Head Latent Attention (MLA) \cite{deepseek2024v2}:} A Key-Value (KV) attention compression technique introduced in DeepSeek-V2~\cite{liu2024deepseek} that reduces memory by jointly encoding keys and values into shared latent vectors, which enable faster inference with minimal performance loss.

\end{itemize}

In this research, we study listed eight attention types in a unified framework for empirical comparison on five different models, including GPT-2 \cite{radford2019language}, ModernBERT \cite{warner2025smarter}, Qwen \cite{bai2023qwen}, Visual Transformer (ViT) \cite{dosovitskiy2021image}, and a Nano-VLM \cite{agarwalla2025nanovlms}.

\section{Experiments}
\label{sec:experiments}

To investigate the performance of various self-attention mechanisms within transformer-based models, we implemented and compared the listed attention modules using the described models. All experiments were conducted in a controlled environment, with the identical settings, to ensure fair comparison. 

\subsection{Experimental Models}
We adopted three representative Transformer-based models as backbones for our experiments: GPT-2 \cite{radford2019language}, ModernBERT \cite{warner2025smarter}, and Qwen \cite{bai2023qwen}. GPT-2 serves as a well-established baseline widely used for experimental and research purposes, while ModernBERT and Qwen represent more recent advances in NLP. For all three models, we followed the same experimental protocol, i.e., the original self-attention implementation was treated as the baseline, and only the self-attention modules were replaced with alternative attention mechanisms in a plug-and-play manner. All other architectural components—such as the number of layers, hidden size, intermediate size, and positional encoding were kept consistent to isolate the impact of attention choice.

Each attention type was implemented as an independent module, designed to be plug-and-play within the transformer block. To conduct our comparison we assume the baseline models self-attention and compared it with Scaled Dot-Product Attention \cite{luong2015effective}, GQA \cite{ainslie2023gqa}, Sliding Window Attention \cite{beltagy2020longformer}, LSH Attention \cite{kitaev2020reformer}, Linear Attention \cite{katharopoulos2020transformers}, Flash Attention version 2 \cite{dao2023flashattention2}, and MLA \cite{deepseek2024v2}. 

We further extended our experiments to two vision-language models, ViT \cite{dosovitskiy2021image} and Nano-VLM \cite{agarwalla2025nanovlms}. For vision-language models, three attention mechanisms were excluded, including Sliding Window Attention, LSH Attention, and Linear Attention. Sliding Window Attention relies on a 1D local window around each token, which does not align well with the flattened 2D image patch sequences used in vision transformers, as sequential neighbors in the 1D sequence may not correspond to spatially meaningful neighbors in the original image. LSH Attention is designed to reduce complexity for long sequences by hashing similar tokens into shared buckets; however, vision transformer patch sequences are typically short (e.g., 196 tokens for a 224$\times$224 image with 16$\times$16 patches), making the hashing overhead impractical relative to the sequence length. Linear Attention was excluded as well.  Linear attention struggles in vision models because removing the softmax operation strips away token-level competition \cite{Fan2024BreakingTL}. This creates a low-rank bottleneck that blurs vital fine-grained local spatial interactions into a smooth, diffused representation \cite{Han2023FLattenT}. Results for these excluded mechanisms are reported as NA in the corresponding figures.

\subsection{Hardware Settings}
The computational experiments were performed on hardware featuring an Intel Core i9 CPU operating at 3.30 GHz, 256 GB of system RAM, and one NVIDIA RTX 4090 GPU, equipped with 24 GB VRAM. The software environment included Ubuntu 20.04 LTS and CUDA toolkit version 12.0.

% \subsection{Experiment's Dataset}

\subsection{Experiment's Dataset}
To evaluate attention mechanisms across different model types, we used three datasets, each selected to match the input requirements of the corresponding architecture.

For the three language models (GPT-2, ModernBERT, and Qwen), we used AllenAI's Tulu-v2-sft-mixture dataset from the Hugging Face Hub\footnote{https://huggingface.co/datasets/allenai/tulu-v2-sft-mixture}. This dataset is designed for instruction tuning of large language models, where each sample contains a structured message list simulating multi-turn human-assistant interactions. These messages were parsed into plain text using a custom text dataset class, tokenized with the GPT2Tokenizer, and padded or truncated to a maximum length of 512 tokens for consistency across samples.

For the Vision Transformer (ViT), we used a sampled subset of the ImageNet training dataset~\cite{deng2009imagenet}, a widely used benchmark for image classification tasks. We sampled up to 50,000 examples from the training split, processed with a batch size of 128.
For the Nano-VLM, we used the VQAv2 configuration of the Cauldron dataset \footnote{https://huggingface.co/datasets/HuggingFaceM4/the\_cauldron}, a multi-modal collection that pairs images with question-answer text. We sampled up to 16,000 examples from the training split. Images were resized to 224$\times$224 pixels and processed alongside their corresponding text through a custom VLM Dataset class, using a batch size of 40.

\subsection{Evaluation Metrics}

Alongside model accuracy, there is increasing emphasis on evaluating computational cost and energy consumption. Studies have shown that inference with large models often results in high GPU usage and power utilization \cite{lacoste2019quantifying, patterson2021carbon, schwartz2020green, strubell2020energy}. Metrics such as FLOPS (floating point operations, a measure of computational complexity), GPU power usage, memory usage, and runtime latency are now commonly used in empirical evaluations \cite{henderson2020towards, schwartz2020green, lu2024impact, khedri2025pruning}. 

In this work, we adopt these metrics to assess the trade-offs of different attention mechanisms from both computational and energy-efficiency perspectives. To this end, we build an open-source monitoring tools to profile CPU/GPU usage, memory allocation, GPU power consumption, and average forward time in real time.

All models were trained for 30 epochs using a batch size of 16, with each epoch consisting of 400 batches. This corresponds to a total of 6,400 training steps per model. Experiments were conducted on a single NVIDIA RTX 4090 GPU, and the training schedule was fixed across all attention variants to ensure fair and consistent comparisons. This setting also reflects practical limitations under resource-constrained environments, such as personal desktop computer.

\section{Results}
In this section we present our empirical results from the seven metrics that we have previously described, on the five models.
% ----------------------------
% Group 1 — Training Time + Inference Time
% ----------------------------
\begin{figure*}[!t]
\hspace{-0.7cm}%
\includegraphics[width=1.05\textwidth,keepaspectratio]{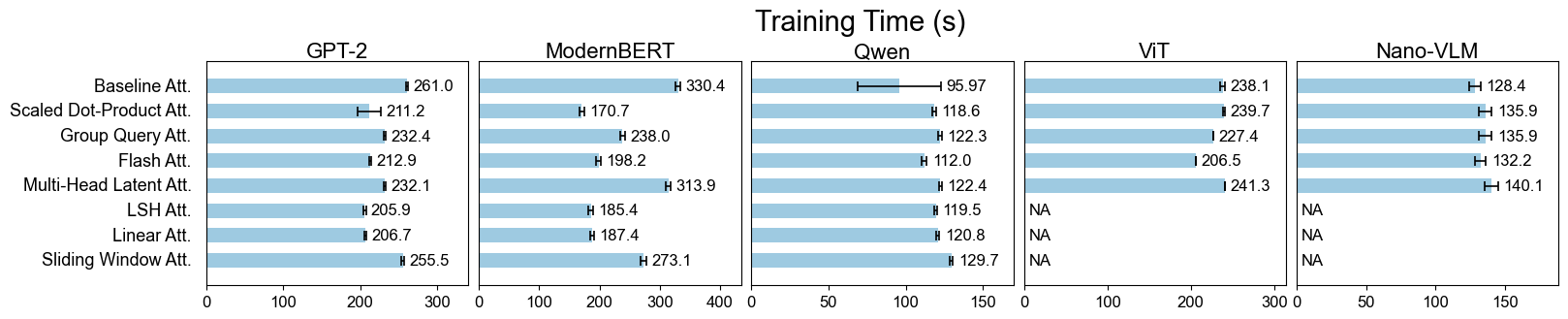}\par\vspace{3mm}
\hspace{-0.7cm}%
\includegraphics[width=1.05\textwidth,keepaspectratio]{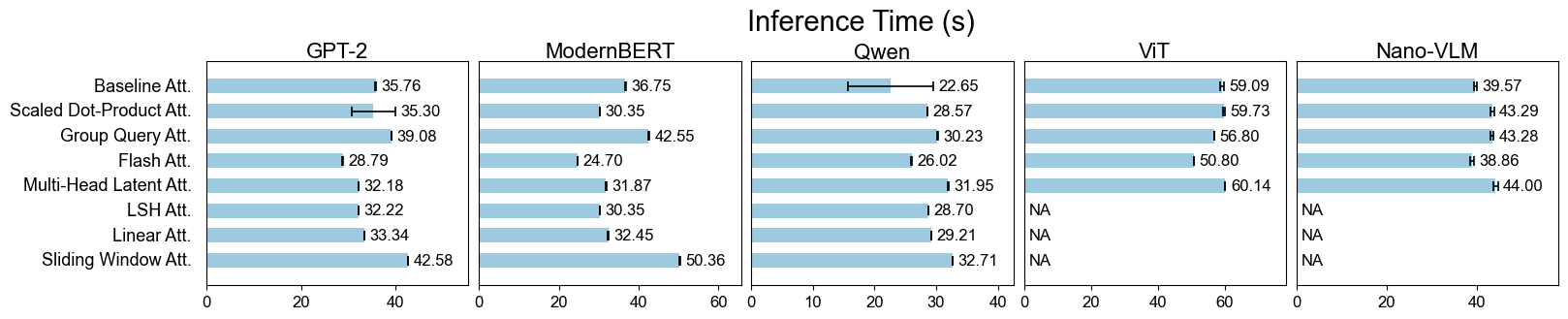}
\caption{Training time per epoch and average inference time per batch across different attention mechanisms for all evaluated models. "Att" is an abbreviation for "Attention". NA indicates mechanisms not evaluated for that model; see Section~\ref{sec:experiments} for details.}
\label{fig:time}
\end{figure*}

% ----------------------------
% Group 2 — GPU Power + GPU Memory + FLOPs
% ----------------------------
\begin{figure*}[!t]
\hspace{-0.7cm}%
\includegraphics[width=1.05\textwidth,keepaspectratio]{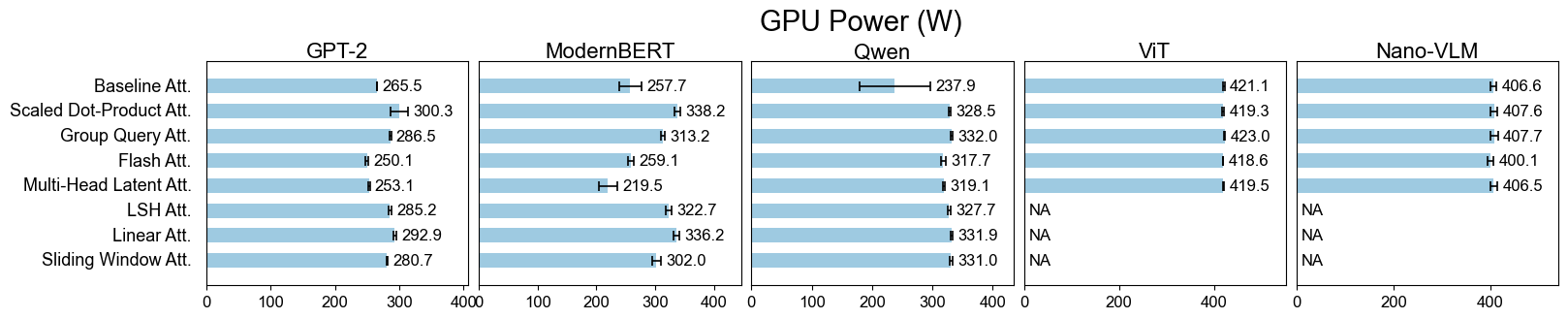}\par\vspace{3mm}
\hspace{-0.7cm}%
\includegraphics[width=1.05\textwidth,keepaspectratio]{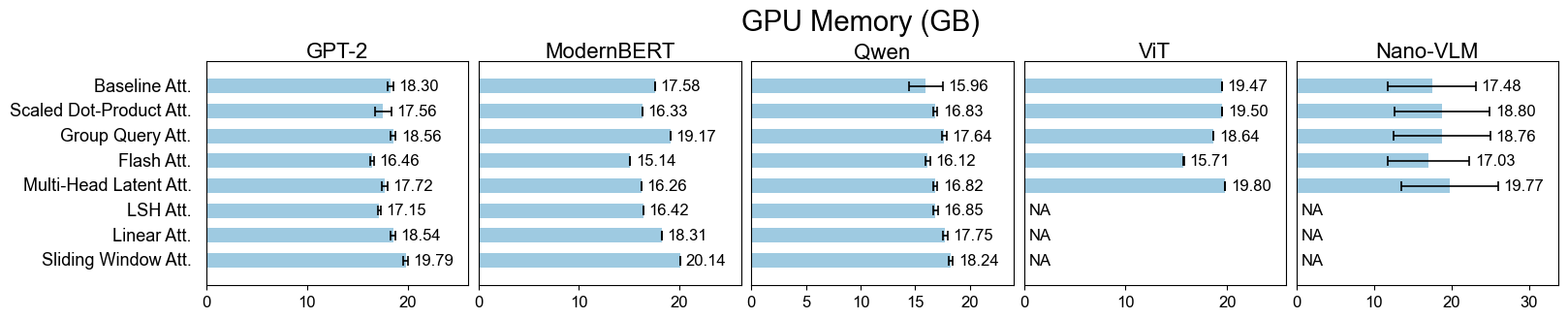}\par\vspace{3mm}
\hspace{-0.7cm}%
\includegraphics[width=1.05\textwidth,keepaspectratio]{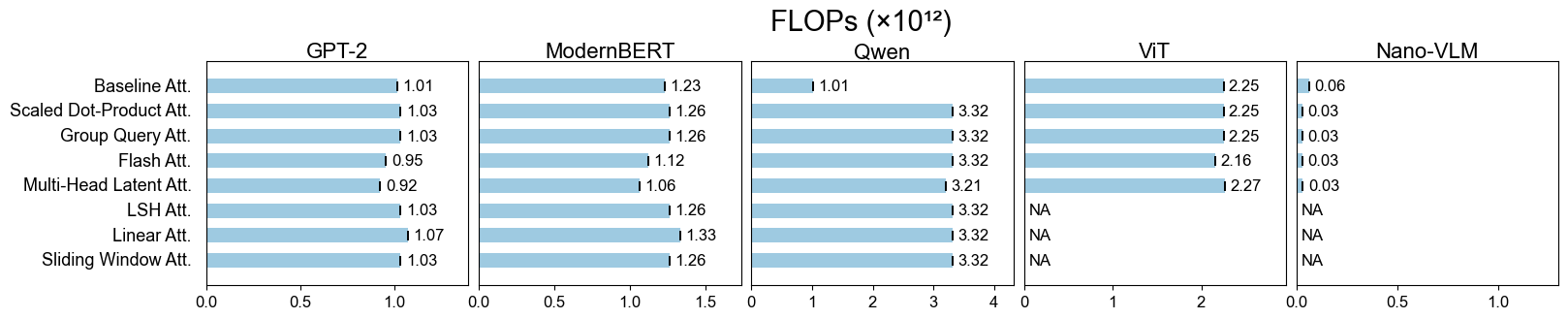}
\caption{GPU power usage, GPU memory consumption, and FLOPS comparison between attention variants across all evaluated models. "Att" is an abbreviation for "Attention". NA indicates mechanisms not evaluated for that model, see Section~\ref{sec:experiments} for details.}
\label{fig:gpu_stats}
\end{figure*}

% ----------------------------
% Group 3 — Total Energy
% ----------------------------
\begin{figure*}[!t]
\hspace{-0.7cm}%
\includegraphics[width=1.05\textwidth,keepaspectratio]{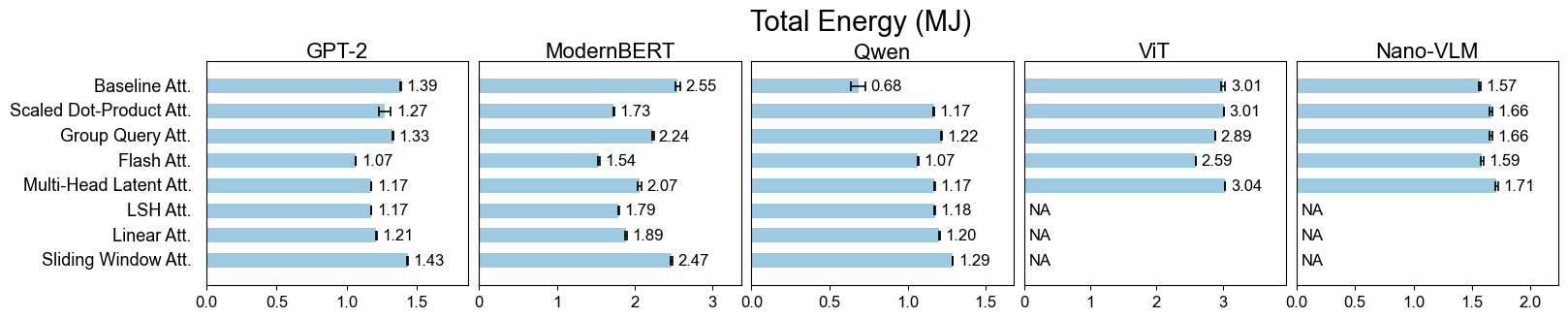}
\caption{Total GPU energy consumption (Watts $\times$ Seconds) for each attention mechanism across all models. "Att" is an abbreviation for "Attention". MJ is an abbreviation for mega-joule. NA indicates that the given attention mechanism is not compatible with the model architecture.}
\label{fig:total_energy}
\end{figure*}

% ----------------------------
% Group 4 — Loss
% ----------------------------
\begin{figure*}[!t]
\hspace{-0.7cm}%
\includegraphics[width=1.05\textwidth,keepaspectratio]{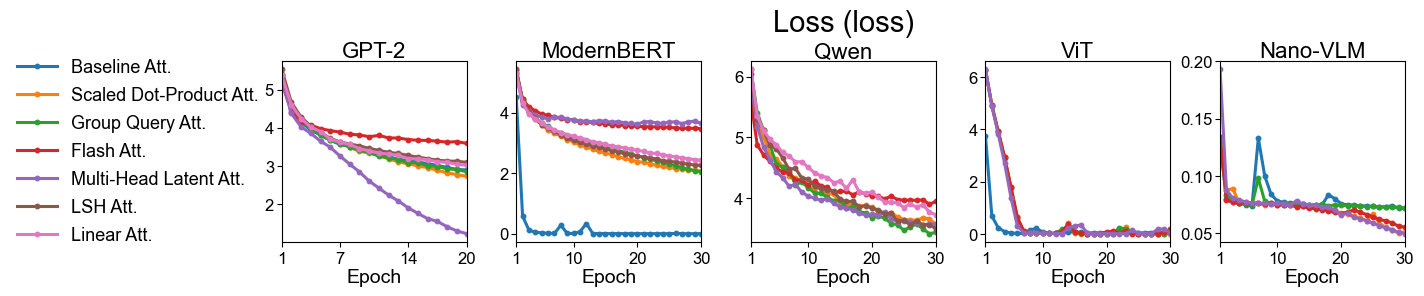}
\caption{Training loss curves across epochs for all evaluated models. }
\label{fig:loss}
\end{figure*}

% ----------------------------
% Group 6 — Memory Usage + Model Size
% ----------------------------
\begin{figure*}[!t]
\hspace{-0.7cm}%
\includegraphics[width=1.05\textwidth,keepaspectratio]{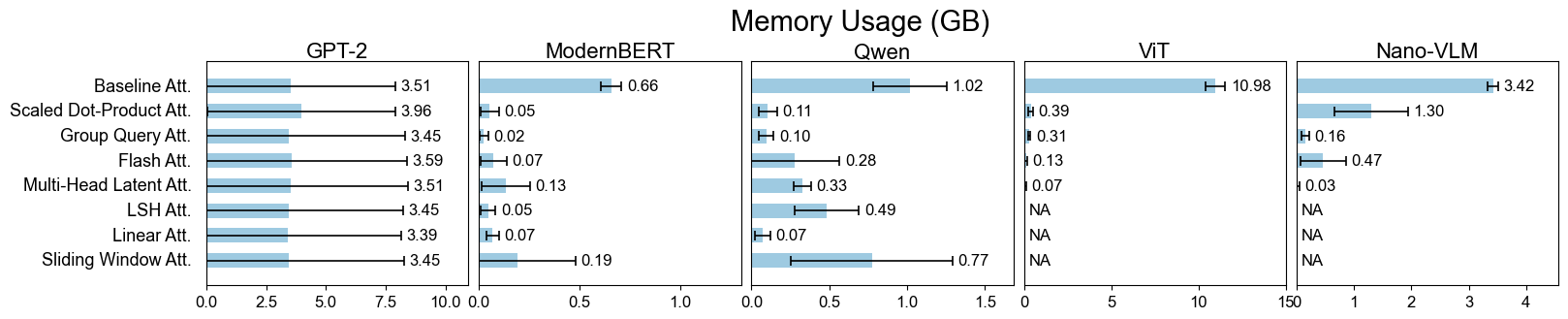}\par\vspace{3mm}
\hspace{-0.7cm}%
\includegraphics[width=1.05\textwidth,keepaspectratio]{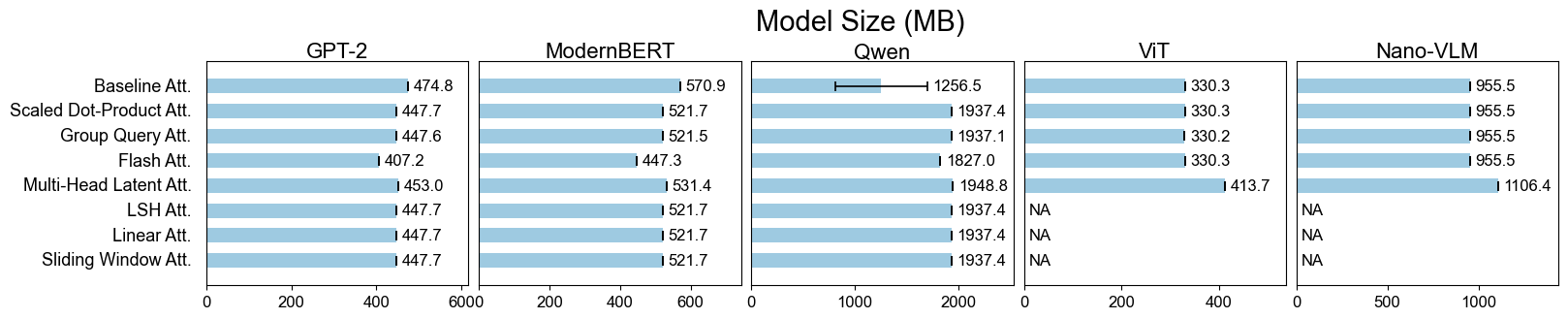}
\caption{Memory usage during training and model size (in MB) for each attention mechanism across all evaluated models. "Att" is an abbreviation for "Attention". NA indicates that the given attention mechanism is not compatible with the model architecture.}
\label{fig:memory_model}
\end{figure*}
% ----------------------------
% Group 5 — GPU Percentage
% ----------------------------
\begin{figure*}[!t]
\hspace{-0.7cm}%
\includegraphics[width=1.05\textwidth,keepaspectratio]{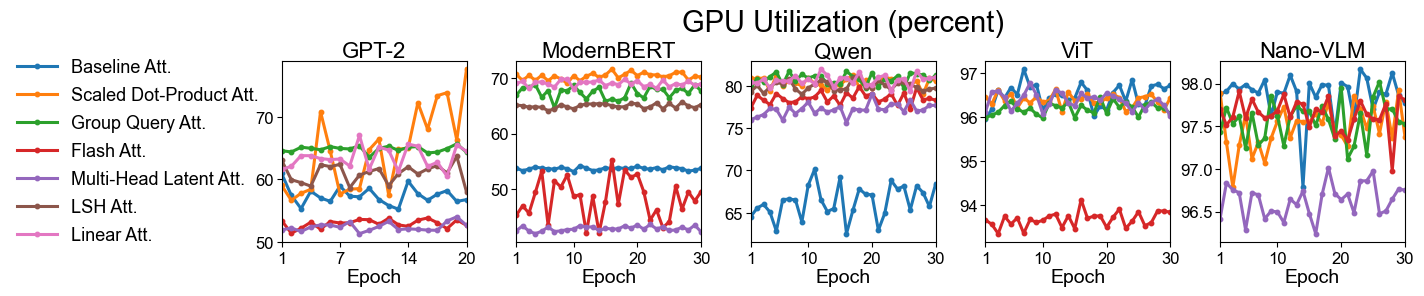}
\caption{GPU utilization percentage during training across all evaluated models. "Att" is an abbreviation for "Attention".}
\label{fig:gpu_util}
\end{figure*}

% ----------------------------
% Group 7 — CPU Percentage
% ----------------------------
\begin{figure*}[!t]
\hspace{-0.7cm}%
\includegraphics[width=1.05\textwidth,keepaspectratio]{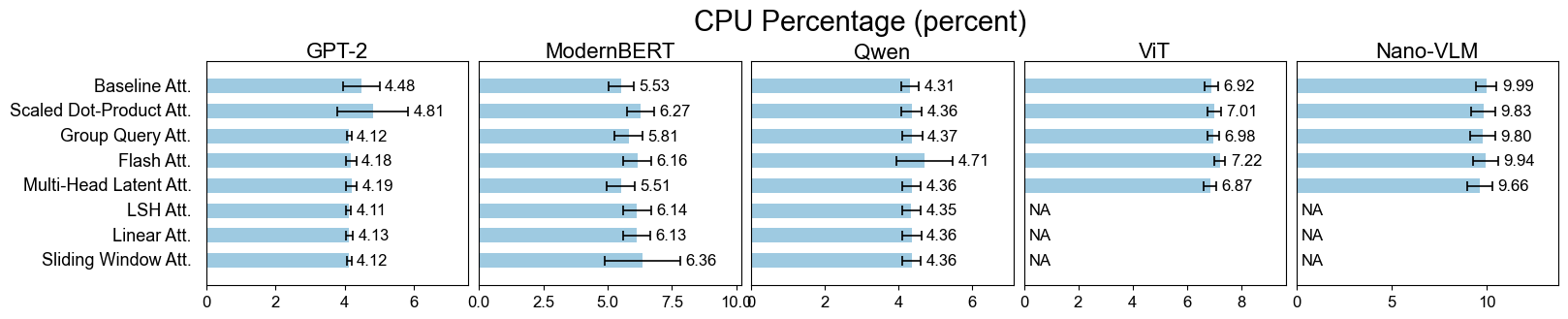}
\caption{CPU utilization percentage during training across different attention mechanisms for all evaluated models. "Att" is an abbreviation for "Attention". NA indicates that the given attention mechanism is not compatible with the model architecture.}
\label{fig:cpu}
\end{figure*}

% We should mention we provide GPT results here and all other results are in appendix. 

\subsection{Training and Inference Times}

Figure~\ref{fig:time} presents the average training time and inference time for all models and attention variants. As it is shown in this figure, LSH Attention, Linear Attention, and Scaled Dot-Product Attention train faster than others on average. Flash Attention is customized for NVIDIA GPU and it is the only Attention function that operates by leveraging the memory architecture of NVIDIA GPUs. Grouped Query Attention and Sliding Window Attention are moderate, while Baseline Attention shows a longer duration for training. Usually, training time is directly associated with energy consumption. Nevertheless, our later results show this phenomenon does not always occur, and there is a correlation between two.

Flash Attention yield the fastest inference times, because it is customized for NVIDIA GPU. Multihead Latent Attention, Linear Attention and LSH Attention are fastest one after Flash Attention. while Sliding Window Attention and Grouped Query Attention are slower.

\subsection {GPU Energy Utilization}

Figure~\ref{fig:gpu_stats} displays GPU power usage during training. Flash Attention and MLA draw less power, due to their optimized GPU memory usage, while Baseline Attention and Scaled Dot-Product Attention remain consistently higher.

Cumulative energy use per attention variant is shown in Figure~\ref{fig:total_energy}. Flash Attention achieves the lowest total energy consumption primarily due to its minimal GPU power usage, while LSH Attention ranks second in energy efficiency, mainly attributed to its fastest training time. MLA demonstrates the third-best overall energy performance, striking a balanced combination of moderate power consumption and training duration.

\subsection{Training Convergence and Resource Usage} 
Based on data presented in Figure~\ref{fig:loss}, we identify that all evaluated models demonstrate normal training convergence, which is a steady decrease in loss values over 20 epochs. This confirms that replacing the Baseline Attention mechanism with some of the listed optimized attention mechanism does not negatively impact the training when properly integrated.

All attention mechanisms maintain comparable model sizes, despite different architectural components, as it has been illustrated in Figure~\ref{fig:memory_model}. This is expected, since the only modified module is the self-attention block.

The FLOPS comparison in Figure~\ref{fig:gpu_stats} shows that MLA and Flash Attention achieve the lowest computational complexity. All other mechanisms, including Linear Attention and Baseline Attention, are within a similar range. 
In terms of GPU memory usage, Figure~\ref{fig:gpu_stats} shows that LSH Attention and Flash Attention are the most memory-efficient, whereas Sliding Window Attention consumes the highest GPU memory during training.
However, Figure~\ref{fig:gpu_util} presents that GPU utilization is very much model dependent and we can not identify any correlation between model type (LLM or VLM) and attention mechanism.

As it has been presented in Figure \ref{fig:cpu} CPU utilization during training is not significantly different compared to other models. This is because the majority of resource utilization overhead during training is on the GPU, while the CPU is mainly utilized at inference time. The same applies to CPU inference, which we did not report here, because the majority of the computational workload for attention mechanisms during inference involves matrix multiplication, which is more efficiently handled by the GPU.

\section{Discussion and Findings}
We lists our findings and lessons we have learned by benchmarking the resource utilization of different attention mechanisms, as follows. 

\textbf{Total energy consumption exhibits distinct dominant factors across LLMs and VLMs.}
As shown in Figure \ref{fig:total_energy}, in GPT-2, Flash Attention achieves the lowest total energy at 1.07 MJ, followed by MLA at 1.17 MJ, while the Baseline consumes 1.39 MJ and Sliding Window Attention reaches 1.43 MJ. In ModernBERT, Flash Attention (1.54 MJ) significantly outperforms the Baseline (2.55 MJ) and Sliding Window Attention (2.47 MJ). These reductions correspond to both lower GPU power (e.g., 259.1–319.3 W range for GPT-2 in Figure~\ref{fig:gpu_stats}) and competitive training time (e.g., 198–232 seconds for Flash/MLA vs. 261 seconds for Baseline in Figure~\ref{fig:time}). However, in Qwen, ViT and Nano-VLM, GPU power remains relatively stable across attention mechanisms. For instance, Qwen’s GPU power ranges narrowly between 317–332 W, while ViT consistently operates around 418–423 W (Figure~\ref{fig:gpu_stats}). As a result, total energy differences in these models are primarily determined by training time rather than instantaneous power. For example, in Qwen, training time varies from 105 seconds to 130 seconds (Figure~\ref{fig:time}), which directly explains the energy variation from 1.07 MJ to 1.29 MJ (Figure~\ref{fig:total_energy}). This pattern indicates that in multimodal architectures, time efficiency dominates energy differentiation.

\textbf{Under a fixed-epoch training setting, training duration linearly scales total energy consumption.}
Since all experiments are conducted with a fixed number of epochs rather than fixed convergence loss, total energy can be approximated as average power multiplied by training time. In GPT-2, Baseline requires 261 seconds per epoch while Flash Attention completes training in 198 seconds, yielding a substantial energy gap despite comparable power levels (Figure~\ref{fig:time}). In Qwen and ViT, where power varies by less than 5\% across mechanisms (Figure~\ref{fig:gpu_stats}), differences in training time (e.g., 206–241 seconds for ViT) translate almost proportionally into total energy differences (2.59–3.04 MJ in Figure~\ref{fig:total_energy}). Loss curves in Figure~\ref{fig:loss} show that convergence behaviors differ across mechanisms; thus, the reported energy values represent consumption under a fixed training budget rather than the minimum energy required to reach equivalent performance. This distinction is critical when interpreting efficiency rankings.

\textbf{Beyond energy, MLA and Flash Attention maintain consistent advantages across computational and deployment metrics.}
In terms of FLOPs (Figure~\ref{fig:gpu_stats}), MLA and Flash Attention achieve lower computational complexity in GPT-2 (0.92–0.95 $\times 10^{12}$ FLOPs) compared to the Baseline (1.01 $\times 10^{12}$). For ModernBERT, FLOPs vary between 1.06–1.63 $\times 10^{12}$, with MLA remaining competitive. Regarding GPU memory (Figure~\ref{fig:gpu_stats}), Flash Attention typically consumes around 15–17 GB in GPT-2 and ModernBERT, lower than Sliding Window Attention (up to 19.8 GB). In inference time (Figure~\ref{fig:time}), Flash Attention and MLA consistently rank among the fastest (e.g., 28–33 seconds in GPT-2 vs. 42 seconds for Sliding Window). CPU utilization remains relatively stable across mechanisms (4–7\% for LLMs, 7–10\% for VLMs; Figure~\ref{fig:cpu}), indicating that efficiency differences are primarily GPU-driven. Taken together, these quantitative results demonstrate that MLA and Flash Attention provide the most balanced trade-off between energy efficiency, computational complexity, and deployment feasibility.

These findings carry important implications for the sustainable model development. In hardware-constrained or energy-sensitive deployment scenarios, mechanisms such as Flash Attention, LSH Attention, and MLA offer higher deployment potential. Sliding Window Attention, despite its structural complexity, do not exhibit clear advantages in terms of energy efficiency. This underscores the importance of considering both resource cost and model accuracy in the design of attention modules.

\section{Conclusion and Future Work}
This study presents a systematic evaluation and benchmarking of various attention mechanisms in terms of resource efficiency during model training. Under a unified architecture and dataset, we quantitatively measured key indicators including training time, GPU power usage, and total energy consumption. The results show that Flash Attention (customized only for NVIDIA GPU) demonstrates the best overall energy efficiency, with the lowest GPU power consumption and competitive training speed. LSH Attention and Linear Attention also achieve low total energy usage due to significantly shorter training durations. MLA ranks third in energy efficiency, benefiting from the second-lowest GPU power consumption, which compensates for its moderate training time and maintains its competitive position. Overall, the most energy-efficient mechanisms tend to strike a strong balance between low power draw and high computational speed.

Future work can expand this study by testing these mechanisms on larger numbers of open models to evaluate their scalability and generalizability. We can also benchmark on more diverse and multi-task datasets to gain insights into performance under complex workloads. 

\bibliographystyle{plain}
\balance
\bibliography{refs}

@article{Morrisonetal2025,
  author  = {Morrison, Jacob and Na, Clara and Fernandez, Jared and Dettmers, Tim and Strubell, Emma and Dodge, Jesse},
  title   = {{Holistically Evaluating the Environmental Impact of Creating Language Models}},
  journal = {arXiv},
  year    = {2025},
  doi     = {10.48550/arxiv.2503.05804}
}

@article{Ebertetal2024,
  author  = {Ebert, Kai and Alder, Nicolas and Herbrich, Ralf and Hacker, Philipp},
  title   = {{AI, Climate, and Regulation: From Data Centers to the AI Act}},
  journal = {arXiv},
  year    = {2024},
  doi     = {10.48550/arxiv.2410.06681}
}

@article{Dooetal2024,
  author  = {Doo, Florence X. and Vosshenrich, Jan and Cook, Tessa S. and Moy, Linda and Almeida, Eduardo P. R. P. and Woolen, Sean A. and Gichoya, Judy Wawira and Heye, Tobias and Hanneman, Kate},
  title   = {{Environmental Sustainability and AI in Radiology: A Double-Edged Sword}},
  journal = {Radiology},
  year    = {2024},
  volume  = {310},
  doi     = {10.1148/radiol.232030}
}

@book{rezabook,
 title={{Machine Learning and Artificial Intelligence: Concepts, Algorithms and Models}},
 author={Rawassizadeh, Reza},
  year={2025},
  isbn={9798992162103}
}

@article{odsearch23,
  title={{ODSearch: Fast and resource efficient on-device natural language search for fitness trackers' Data}},
  author={Rawassizadeh, Reza and Rong, Yi},
  journal={Proceedings of the ACM on Interactive, Mobile, Wearable and Ubiquitous Technologies},
  volume={6},
  number={4},
  pages={1--25},
  year={2023},
  publisher={ACM New York, NY, USA}
}

@article{ejimuda2025,
  title={{Analyzing the Resource Utilization of Lambda Functions on Mobile Devices: Case Studies on Kotlin and Swift}},
  author={Ejimuda, Chibundom and Longhitano, Gaston and Rawassizadeh, Reza},
  journal={IEEE Pervasive Computing},
  volume={24},
  number={2},
  pages={48--51},
  year={2025},
  publisher={IEEE}
}

@article{bux2025critical,
  title={{A Critical Analysis of Global Warming Potential of Data Centers in the Digital Era}},
  author={Bux, Christian and Rana, Roberto Leonardo and Lombardi, Mariarosaria and Giungato, Pasquale and Tricase, Caterina},
  journal={The International Journal of Life Cycle Assessment},
  volume={30},
  number={11},
  pages={2390--2402},
  year={2025},
  publisher={Springer}
}

@article{wen2025grades,
  title={{GradES: Significantly Faster Training in Transformers with Gradient-Based Early Stopping}},
  author={Wen, Qifu and Zeng, Xi and Zhou, Zihan and Liu, Shuaijun and Hosseinzadeh, Mehdi and Su, Ningxin and Rawassizadeh, Reza},
  journal={arXiv preprint arXiv:2509.01842},
  year={2025}
}

@article{liu2024deepseek,
  title={Deepseek-v2: A strong, economical, and efficient mixture-of-experts language model},
  author={Liu, Aixin and Feng, Bei and Wang, Bin and Wang, Bingxuan and Liu, Bo and Zhao, Chenggang and Dengr, Chengqi and Ruan, Chong and Dai, Damai and Guo, Daya and others},
  journal={arXiv preprint arXiv:2405.04434},
  year={2024}
}

@article{vaswani2017attention,
  title={Attention is all you need},
  author={Vaswani, Ashish and Shazeer, Noam and Parmar, Niki and Uszkoreit, Jakob and Jones, Llion and Gomez, Aidan N and Kaiser, {\L}ukasz and Polosukhin, Illia},
  journal={Advances in neural information processing systems},
  volume={30},
  year={2017}
}

@article{luong2015effective,
  title={Effective approaches to attention-based neural machine translation},
  author={Luong, Minh-Thang and Pham, Hieu and Manning, Christopher D},
  journal={arXiv preprint arXiv:1508.04025},
  year={2015}
}

@article{dao2023flashattention2,
  title={FlashAttention-2: Faster Attention with Better Parallelism and Work Partitioning},
  author={Dao, Tri and Schuster, Lev and Fu, Haokun and Bailis, Peter and Liang, Percy},
  journal={arXiv preprint arXiv:2307.08691},
  year={2023}
}

@article{li2025minimax,
  title={{Minimax-01: Scaling foundation models with lightning attention}},
  author={Li, Aonian and Gong, Bangwei and Yang, Bo and Shan, Boji and Liu, Chang and Zhu, Cheng and Zhang, Chunhao and Guo, Congchao and Chen, Da and Li, Dong and others},
  journal={arXiv preprint arXiv:2501.08313},
  year={2025}
}

@inproceedings{ainslie2023gqa,
  title={GQA: Training Generalized Multi-Query Transformer Models from Multi-Head Checkpoints},
  author={Ainslie, Joshua and Lee-Thorp, James and de Jong, Michiel and Zemlyanskiy, Yury and Lebrón, Federico},
  booktitle={Proceedings of the 2023 Conference on Empirical Methods in Natural Language Processing},
  year={2023}
}

@article{deepseek2024v2,
  title={DeepSeek-V2: A Strong, Economical, and Efficient Mixture-of-Experts Language Model},
  author={DeepSeek-AI},
  journal={arXiv preprint arXiv:2405.04434},
  year={2024}
}

@article{radford2019language,
  title={Language models are unsupervised multitask learners},
  author={Radford, Alec and Wu, Jeffrey and Child, Rewon and Luan, David and Amodei, Dario and Sutskever, Ilya and others},
  journal={OpenAI blog},
  volume={1},
  number={8},
  pages={9},
  year={2019}
}

@article{kitaev2020reformer,
  title={Reformer: The efficient transformer},
  author={Kitaev, Nikita and Kaiser, {\L}ukasz and Levskaya, Anselm},
  journal={arXiv preprint arXiv:2001.04451},
  year={2020}
}

@article{kaplan2020scaling,
  title={Scaling laws for neural language models},
  author={Kaplan, Jared and McCandlish, Sam and Henighan, Tom and Brown, Tom B and Chess, Benjamin and Child, Rewon and Gray, Scott and Radford, Alec and Wu, Jeffrey and Amodei, Dario},
  journal={arXiv preprint arXiv:2001.08361},
  year={2020}
}

@article{brown2020language,
  title={Language models are few-shot learners},
  author={Brown, Tom and Mann, Benjamin and Ryder, Nick and Subbiah, Melanie and Kaplan, Jared D and Dhariwal, Prafulla and Neelakantan, Arvind and Shyam, Pranav and Sastry, Girish and Askell, Amanda and others},
  journal={Advances in neural information processing systems},
  volume={33},
  pages={1877--1901},
  year={2020}
}

@article{zhao2023survey,
  title={A survey of large language models},
  author={Zhao, Wayne Xin and Zhou, Kun and Li, Junyi and Tang, Tianyi and Wang, Xiaolei and Hou, Yupeng and Min, Yingqian and Zhang, Beichen and Zhang, Junjie and Dong, Zican and others},
  journal={arXiv preprint arXiv:2303.18223},
  volume={1},
  number={2},
  year={2023}
}

@article{tay2022efficient,
  title={Efficient transformers: A survey},
  author={Tay, Yi and Dehghani, Mostafa and Bahri, Dara and Metzler, Donald},
  journal={ACM Computing Surveys},
  volume={55},
  number={6},
  pages={1--28},
  year={2022},
  publisher={ACM New York, NY}
}

@inproceedings{strubell2020energy,
  title={Energy and policy considerations for modern deep learning research},
  author={Strubell, Emma and Ganesh, Ananya and McCallum, Andrew},
  booktitle={Proceedings of the AAAI conference on artificial intelligence},
  volume={34},
  number={09},
  pages={13693--13696},
  year={2020}
}

@article{schwartz2020green,
  title={Green ai},
  author={Schwartz, Roy and Dodge, Jesse and Smith, Noah A and Etzioni, Oren},
  journal={Communications of the ACM},
  volume={63},
  number={12},
  pages={54--63},
  year={2020},
  publisher={ACM New York, NY, USA}
}

@article{patterson2021carbon,
  title={Carbon emissions and large neural network training},
  author={Patterson, David and Gonzalez, Joseph and Le, Quoc and Liang, Chen and Munguia, Lluis-Miquel and Rothchild, Daniel and So, David and Texier, Maud and Dean, Jeff},
  journal={arXiv preprint arXiv:2104.10350},
  year={2021}
}

@article{lacoste2019quantifying,
  title={Quantifying the carbon emissions of machine learning},
  author={Lacoste, Alexandre and Luccioni, Alexandra and Schmidt, Victor and Dandres, Thomas},
  journal={arXiv preprint arXiv:1910.09700},
  year={2019}
}

@article{touvron2023llama,
  title={Llama 2: Open foundation and fine-tuned chat models},
  author={Touvron, Hugo and Martin, Louis and Stone, Kevin and Albert, Peter and Almahairi, Amjad and Babaei, Yasmine and Bashlykov, Nikolay and Batra, Soumya and Bhargava, Prajjwal and Bhosale, Shruti and others},
  journal={arXiv preprint arXiv:2307.09288},
  year={2023}
}

@article{beltagy2020longformer,
  title={Longformer: The long-document transformer},
  author={Beltagy, Iz and Peters, Matthew E and Cohan, Arman},
  journal={arXiv preprint arXiv:2004.05150},
  year={2020}
}

@inproceedings{katharopoulos2020transformers,
  title={Transformers are RNNs: Fast autoregressive transformers with linear attention},
  author={Katharopoulos, Angelos and Vyas, Apoorv and Pappas, Nikolaos and Fleuret, Fran{\c{c}}ois},
  booktitle={International conference on machine learning},
  pages={5156--5165},
  year={2020},
  organization={PMLR}
}

@article{cai2024survey,
  title={A survey on mixture of experts},
  author={Cai, Weilin and Jiang, Juyong and Wang, Fan and Tang, Jing and Kim, Sunghun and Huang, Jiayi},
  journal={arXiv preprint arXiv:2407.06204},
  year={2024}
}

@article{hu2022lora,
  title={Lora: Low-rank adaptation of large language models.},
  author={Hu, Edward J and Shen, Yelong and Wallis, Phillip and Allen-Zhu, Zeyuan and Li, Yuanzhi and Wang, Shean and Wang, Lu and Chen, Weizhu and others},
  journal={ICLR},
  volume={1},
  number={2},
  pages={3},
  year={2022}
}

@article{chaplot2023albert,
  title={Albert q. jiang, alexandre sablayrolles, arthur mensch, chris bamford, devendra singh chaplot, diego de las casas, florian bressand, gianna lengyel, guillaume lample, lucile saulnier, l{\'e}lio renard lavaud, marie-anne lachaux, pierre stock, teven le scao, thibaut lavril, thomas wang, timoth{\'e}e lacroix, william el sayed},
  author={Chaplot, Devendra Singh},
  journal={arXiv preprint arXiv:2310.06825},
  year={2023}
}

@inproceedings{charikar2002,
  title={{Similarity Estimation Techniques from Rounding Algorithms}},
  author={Charikar, Moses S},
  booktitle={Proceedings of the thiry-fourth annual ACM symposium on Theory of computing},
  pages={380--388},
  year={2002}
}

@inproceedings{warner2025smarter,
  title={{Smarter, better, faster, longer: A modern bidirectional encoder for fast, memory efficient, and long context finetuning and inference}},
  author={Warner, Benjamin and Chaffin, Antoine and Clavi{\'e}, Benjamin and Weller, Orion and Hallstr{\"o}m, Oskar and Taghadouini, Said and Gallagher, Alexis and Biswas, Raja and Ladhak, Faisal and Aarsen, Tom and others},
  booktitle={Proceedings of the 63rd Annual Meeting of the Association for Computational Linguistics (Volume 1: Long Papers)},
  pages={2526--2547},
  year={2025}
}

@article{bai2023qwen,
  title={{Qwen Technical Report}},
  author={Bai, Jinze and Bai, Shuai and Chu, Yunfei and Cui, Zeyu and Dang, Kai and Deng, Xiaodong and Fan, Yang and Ge, Wenbin and Han, Yu and Huang, Fei and others},
  journal={arXiv preprint arXiv:2309.16609},
  year={2023}
}

@inproceedings{dosovitskiy2021image,
  title={{An Image is Worth 16x16 Words: Transformers for Image Recognition at Scale}},
  author={Dosovitskiy, Alexey and Beyer, Lucas and Kolesnikov, Alexander and Weissenborn, Dirk and Zhai, Xiaohua and Unterthiner, Thomas and Dehghani, Mostafa and Minderer, Matthias and Heigold, Georg and Gelly, Sylvain and Uszkoreit, Jakob and Houlsby, Neil},
  booktitle={International Conference on Learning Representations (ICLR)},
  year={2021}
}

@article{agarwalla2025nanovlms,
  title={{NanoVLMs: How small can we go and still make coherent Vision Language Models?}},
  author={Agarwalla, Mukund and Kumar, Himanshu and Dandekar, Raj and Dandekar, Rajat and Panat, Sreedath},
  journal={arXiv preprint arXiv:2502.07838},
  year={2025}
}

@article{henderson2020towards,
  title     = {Towards the Systematic Reporting of the Energy and Carbon Footprints of Machine Learning},
  author    = {Henderson, Peter and Hu, Jieru and Romoff, Joshua and Brunskill, Emma and Jurafsky, Dan and Pineau, Joelle},
  journal   = {Journal of Machine Learning Research},
  volume    = {21},
  number    = {248},
  pages     = {1--43},
  year      = {2020}
}

@inproceedings{deng2009imagenet,
  title     = {ImageNet: A Large-Scale Hierarchical Image Database},
  author    = {Deng, Jia and Dong, Wei and Socher, Richard and Li, Li-Jia and Li, Kai and Fei-Fei, Li},
  booktitle = {2009 IEEE Conference on Computer Vision and Pattern Recognition},
  pages     = {248--255},
  year      = {2009},
  publisher = {IEEE},
  doi       = {10.1109/CVPR.2009.5206848}
}

@article{Fan2024BreakingTL,
  author  = {Fan, Qihang and Huang, Huaibo and He, Ran},
  title   = {{Breaking the Low-Rank Dilemma of Linear Attention}},
  journal = {arXiv},
  year    = {2024},
  doi     = {10.48550/arxiv.2411.07635}
}

@inproceedings{Han2023FLattenT,
  author    = {Han, Dongchen and Pan, Xuran and Han, Yizeng and Song, Shiji and Huang, Gao},
  title     = {{FLatten Transformer: Vision Transformer using Focused Linear Attention}},
  booktitle = {2023 IEEE/CVF International Conference on Computer Vision (ICCV)},
  year      = {2023},
  pages     = {5938--5948},
  doi       = {10.1109/iccv51070.2023.00548}
}

@article{lu2024impact,
  title   = {{The Impact of Quantization and Pruning on Deep Reinforcement Learning Models}},
  author  = {Lu, Heng and Alemi, Mehdi and Rawassizadeh, Reza},
  journal = {arXiv preprint arXiv:2407.04803},
  year    = {2024},
  doi     = {10.48550/arXiv.2407.04803},
  url     = {https://arxiv.org/abs/2407.04803}
}

@article{khedri2025pruning,
  title   = {Pruning and Quantization Impact on Graph Neural Networks},
  author  = {Khedri, Khatoon and Rawassizadeh, Reza and Wen, Qifu and Hosseinzadeh, Mehdi},
  journal = {arXiv preprint arXiv:2510.22058},
  year    = {2025},
  doi     = {10.48550/arXiv.2510.22058}
}
% ============================================================
% Appendix: Grouped by Metric (All Models Together)
% Special layout for first two LINE-PLOT groups:
%   Page1: A.1 -> 4 plots (2x2)
%   Page2: A.1 last plot (left) + A.2 first 2 plots (side-by-side)
%   Page3: A.2 remaining 3 plots (2 top + 1 left-bottom)
% From A.3 onward: one page per metric (5 plots: 2x2 + last-left)
% ============================================================

\end{document}